\documentclass[runningheads,a4paper]{llncs}

\usepackage{times}
\usepackage{amssymb}
\usepackage{graphicx}
\usepackage{url}

\newcommand\citep{\cite}
\usepackage{todonotes}
\usepackage{tikz}
\usepackage{svg}

\usepackage{microtype}

\usepackage{graphicx}

\usepackage{subfig}

\usepackage{multirow}

\begin{document}

\title{Speech2Phone: A Novel and Efficient Method for Training Speaker Recognition Models}
\titlerunning{Speech2Phone: A Efficient Method for Training Speaker Recognition Models }

\author{
Edresson Casanova\inst{1}\thanks{~Corresponding author: edresson@usp.br}, 
Arnaldo Candido Junior\inst{2},
Christopher Shulby\inst{3},
Frederico Santos de Oliveira\inst{4},
Lucas Rafael Stefanel Gris\inst{2},
Hamilton Pereira da Silva\inst{2},
Sandra Aluisio\inst{1},
 Moacir Antonelli Ponti\inst{1}
}

\authorrunning{Casanova et al.}

\institute{University of São Paulo, São Carlos, Brazil
\and
Federal University of Technology - Paraná, Medianeira, Brazil
\and
    DefinedCrowd Corp., Seattle - WA, USA
\and
  Federal University of Goias, Goiânia, Brazil
}


%
%
%
\maketitle              

%
%
%
%

\begin{abstract}
In this paper we present an efficient method for training models for speaker recognition using small or under-resourced datasets. This method requires less data than other SOTA (State-Of-The-Art) methods, e.g. the Angular Prototypical and GE2E loss functions, while achieving similar results to those methods. This is done using the knowledge of the reconstruction of a phoneme in the speaker's voice. For this purpose, a new dataset was built, composed of 40 male speakers, who read sentences in Portuguese, totaling approximately 3h. We compare the three best architectures trained using our method to select the best one, which is the one with a shallow architecture. Then, we compared this model with the SOTA method for the speaker recognition task: the Fast ResNet--34 trained with approximately 2,000 hours, using the loss functions Angular Prototypical and GE2E. Three experiments were carried out with datasets in different languages. Among these three experiments, our model achieved the second best result in two experiments and the best result in one of them. This highlights the importance of our method, which proved to be a great competitor to SOTA speaker recognition models, with 500x less data and a simpler approach. 


\end{abstract}

\section{Introduction}

Voice recognition is widely used in many applications, such as intelligent personal assistants \citep{siri}, telephone-banking systems \citep{2001voice}, automatic question response \citep{watson}, among others. In several of these applications, it is useful to identify the speaker, as is the case in voice-enabled authentication and meeting loggers. \textcolor{black}{Speaker verification can be done in two scenarios: open-set and closed-set. In both scenarios, the objective is to verify that two audio samples belong to the same speaker. However, in the closed-set scenario, the verification is restricted only to speakers seen during the training of the models. On the other hand, in the open-set scenario, verification occurs with speakers not seen in model training \cite{ertacs2011fundamentals,kekre2013closed}. The verification of speakers in an open-set scenario is especially desired in applications such as meeting loggers, since in these applications speakers can be added frequently, thus, the use of closed-set models would imply the retraining of the model after the insertion of new speakers.}

The first works to use deep neural networks in speaker recognition in an open-set scenario learned speaker embeddings were \citep{snyder2018x,snyder2017deep}, using the Softmax function. Although the Softmax classifier can learn different embeddings for different speakers, it is not discriminatory enough \citep{chung2020in}. To work around this problem, models trained with softmax were combined with back-ends built on Probabilistic Linear Discriminant Analysis (PLDA) \citep{ioffe2006probabilistic} to generate scoring functions \citep{ramoji2020pairwise,snyder2018x}. 
On the other hand, Softmax Angular \citep{liu2017sphereface} was proposed and it uses cosine similarity as a logit entry for the softmax layer, and it proved to be superior to the use of softmax only. 

Thereafter, Additive Margins in Softmax (AM-Softmax) \citep{wang2018additive} was proposed to increase inter-class variance by introducing a cosine margin penalty in the target logit. However, according to \citep{chung2020in} training with AM-Softmax and AAM-Softmax \citep{deng2019arcface} proved to be a challenge, as they are sensitive to scale and margin value in the loss function.
The use of Contrastive Loss \citep{chopra2005learning} and Triplet Loss \citep{schroff2015facenet,bredin2017} also achieved promising results in speaker recognition, but these methods require careful choice of pairs or triplets, which costs time and can interfere with performance \citep{chung2020in}.
Finally, the use of Prototypical networks \citep{wang2019centroid} for speaker recognition was proposed. Prototypical networks seek to learn a metric space in which the classification of open-sets of speakers can be performed by calculating distances to prototypical representations of each class. Generalized end-to-end loss (GE2E) \citep{wan2018generalized} and Angular Prototypical (AngleProto) \citep{chung2020in} follow the same principle and achieved state-of-the-art (SOTA) results recently in speaker recognition. \textcolor{black}{Parallel with this work, \cite{heo2020clova} proposed the use of the AngleProto loss function in conjunction with Softmax, presenting a result superior to the use of AngleProto only. The authors proposed a new architecture presenting SOTA results.}

In this work, we propose a new method for training speaker recognition models, called Speech2Phone. This method was trained with approximately 3.5 hours of speech and surpassed a model trained with 2.000 hours of speech using the GE2E loss function. Our method is based on the reconstruction of the pronunciation of a specific phoneme and has shown promise in scenarios with few available resources. In addition, the simplicity of its architecture makes our method suitable for real-time applications with low processing power.


Finally, to simplify the reproduction of this work, Python code and download links for the datasets used to reproduce all experiments are publicly available on the Github repository \footnote{\url{https://github.com/Edresson/Speech2Phone}}

This work is organized as follows. Section \ref{sec:method} details the datasets used as well as the preprocessing performed to attend the proposed experiments. Section \ref{sec:speech2phone-propo} presents the Speech2Phone method and experiments carried out to find the best model trained with this method for the identification of speakers in an open-set scenario.  Section \ref{sec:speaker-verification-experiments} compares the best model trained with the Speech2Phone method with the state of the art in the literature. Finally, Section \ref{sec:conc} shows the conclusions of this paper.

\section{Datasets and pre-processing} \label{sec:method}

Section \ref{sec:method:base} presents the datasets used, as well as describes a new dataset created to attend the needs of our experiments. 
Sections \ref{sec:prepro-Speech2Phone} and \ref{sec:prepro-veri} details the pre-processing performed on the datasets to allow the execution of all the proposed experiments. 


\subsection{Audio datasets}\label{sec:method:base}


The VCTK \citep{veaux2016superseded} is an English language dataset with a total of 109 speakers. During its creation, the 109 speakers spoke approximately 400 sentences. The same phrases are spoken by all speakers. The dataset has approximately 44 hours of speech and is sampled at 48KHz.


Common Voice (CV) \citep{ardila2019common} is a massively multilingual transcribed speech dataset for research and development of speech technology. CV has 54 subsets and each of these sets have data from a language, currently the dataset has a total of 5,671 hours. \textcolor{black}{In this work, we use version 5.2 of the corpus.}


\textcolor{black}{To train our method, it was necessary to build a specific dataset, which we call the Speech2Phone dataset}. This dataset includes 40 male speakers, aged between 20 and 50 years. The dataset includes only Portuguese utterances, because that is the native language of the speakers. 
We chose to focus only on male speakers, because during the collection phase of our dataset we were able to collect only voices from 5 female speakers.
To collect the data, each speaker was given a phonetically balanced text, according to the work of \citep{seara1994estudo}, which was comprised of 149 words. The reading time ranged from 42 to 95 seconds, totaling approximately 43 minutes of speech.

Additionally, we asked each speaker to say the phoneme /a/ for approximately three seconds. The central second of each capture was extracted and then used as expected output for the embedding models. The phoneme /a/ was chosen because it is simple to articulate and very frequent in the Portuguese language. The Speech2Phone dataset is publicly available on the Github repository\footnote{\url{https://github.com/Edresson/Speech2Phone}}

\subsection{Preprocessing of the Speech2Phone dataset}\label{sec:prepro-Speech2Phone}

To preprocess the Speech2Phone dataset we extracted five-second speech segments from the original audio length. The five-second window was defined after preliminary experiments, varying the input duration. In order to maximize the number of speech segments, we used the overlapping technique, in which the window was shifted one second each time and an instance was extracted during the total audio duration. The main dataset resulted in 2,394 speech segments totaling 3 hours and 23 minutes of speech. The next step was to divide the dataset into smaller sets to attend  the needs of each proposed experiment. Therefore, the original dataset was divided into four subsets.
\textcolor{black}{The Partitions $A_1$ and $B_1$ each have 20 different speakers and have approximately 1,097 samples each. Partitions $A_2$ and $B_2$ have approximately 100 speech segments each and have, respectively, speakers from the $A_1$ and $B_1$ partition. Thus, $A$ partitions do not have speakers in common with  $B$ partitions.}



\subsection{Pre-processing of speaker verification datasets}\label{sec:prepro-veri}


We preprocessed the VCTK and CV datasets in order to use them for speaker verification. For the VCTK we chose to use the entire dataset and for CV we used the test subsets of the dataset in Portuguese (PT), Spanish (ES) and Chinese spoken in China (ZH).

The VCTK dataset has, in many of its samples, long initial and final silences. In order to ensure that this feature does not affect our analysis, we chose to remove these silences. So, we applied Voice Activity Detection (VAD) using the Python implementation of the Webrtcvad toolkit\footnote{https://github.com/wiseman/py-webrtcvad}.


We used VCTK and CV to test our models. The datasets were used to build audio pairs. The positive class is composed from audio pairs from the same speaker, while negative class has pairs from different speakers. In this scenario, it is possible to build more examples from the negative class. To avoid class imbalance issues, we defined the maximum number of negative pairs analyzed as the number of positive samples divided by the number of speakers. We also removed speakers with less than two samples.

Table \ref{tab:preproveri} shows the number of speakers, language and number of positive and negative speech segments of the datasets used to verify the speaker in our experiments. The Python code used for the preprocessing of the dataset, as well as the link to download the versions of the VCTK and CV datasets used are available in the Github repository\footnote{\url{https://github.com/Edresson/Speech2Phone/tree/master/Paper/EER-Experiments}}.



\begin{table}[ht]

\centering
\caption{Preprocessed Speaker Verification datasets}
\label{tab:preproveri}
\begin{tabular}{l|l|r|r|r}
\hline
\textbf{Dataset}                                                         & \textbf{Language} & \textbf{Nº Speakers} & \textbf{Nº Pos. Samples}  & \textbf{Nº Neg. Samples}  \\\hline
\multirow{3}{*}{\begin{tabular}[c]{@{}c@{}}Common \\ Voice\end{tabular}} & PT                &        525      & 25,846 &   25,847                  \\
                                                                         & ES                &        4,167   & 19,355 &   19,356                      \\
                                                                         & ZH                &       1,968    & 14,656 &   14,657                           \\
VCTK                                                                     & EN                &         109     & 9,084,638 &   9,001,368               \\

\hline
\end{tabular}

\end{table}







\section{Proposed Method: Speech2Phone} \label{sec:speech2phone-propo}

Section \ref{sec:method:Speech2Phone} presents the experiments carried out to choose the best model trained with the Speech2Phone method to identify speakers in the open-set scenario. On the other hand, section \ref{sec:results:experiuments:Speech2Phone} presents the results of Speech2Phone experiments.

\subsection{Speech2Phone Experiments}\label{sec:method:Speech2Phone}

The goal of open-set models is to be speaker independent; additionally, a desirable feature is to be multilingual and text independent. In pursuit of these goals, we propose that 
the neural network training uses five second speech fragments as input and, as expected output, the reconstruction of a second of a simple phoneme (/a/ in our experiments). As the phoneme sounds differ according to each speaker, a good reconstruction would allow the model to distinguish between speakers. Focusing on a single phoneme allows for dimensionality reduction in the embedding layer. 

In the Speech2Phone experiments, the models input and expected output is represented in the form of Mel Frequency Cepstral Coeficients (MFCCs) \citep{logan2000mel}. We extract MFCCs using the Librosa~\citep{mcfee2015} library. The default sampling rate (22KHz) was used. \textcolor{black}{We chose to empirically extract 13 MFCCs}. Windowed frames were used as defined by the default parameters in Librosa 0.6, namely, a 512 Hop Length and 2,048 as the window length for the Fast Fourier Transform \citep{nussbaumer1981fast}.  In addition, as the models must reconstruct an MFCC segment they were induced using Mean Square Error (MSE) \citep{goodfellow2016}.

Several models using the Speech2Phone method, and consequently the Speech2Phone dataset, have been tested. We report the three most interesting ones in this section. To evaluate these experiments, we used the accuracy of the speaker identification.

\textcolor{black}{To calculate the accuracy, we randomly chose and entered an embedding of a sample of each of the speakers in a database}. For the calculation it is necessary to search the extracted embedding in an embeddings database. This is done by running the KNN \citep{mitchell2013artificial} algorithm with $k=1$ and using the Euclidean distance. If the embedding of a speaker has a Euclidean distance closer to its embedding registered in the database than the embedding of all other speakers, this will be counted as a hit, otherwise it will be counted as an error.

\textcolor{black}{In the experiments, we compared the performance of the open-set models in a closed-set scenario. All of these experiments were trained with the Speech2Phone dataset $A_1$ subset, described previously in Section \ref{sec:prepro-Speech2Phone}. Therefore, the models were trained with only 1,037 speech segments of 20 different speakers, approximately 1.5 hours of speech.}

Tensorflow \citep{abadi2016} and TFlearn \citep{tflearn2016} were used to generate the neural networks for all Speech2Phone Experiments.
All models were trained using the Adam Optimizer \citep{kingma2014adam}. The convolutional layers in all experiments have a stride of 1. Hyperparameters used to train each model are presented in Table \ref{tab:hyper}. 


\begin{table}[ht]

\centering
\caption{Experiments hyperparameters}
\label{tab:hyper}
\begin{tabular}{l|r|r|r}
\hline
\textbf{Experiment}   & \textbf{Epochs} & \textbf{Learning Rate} & \textbf{Batch Size } \\ \hline
1 (Dense)   & 1,000  & 0.0007       & 128          \\
2 (Conv)    & 10     & 0.0010       & 16           \\
3 (Recurrent)  & 100    & 0.0050       & 256          \\
\hline
\end{tabular}

\end{table}

\begin{figure}[]
  \centering
  \begin{minipage}[b]{0.49\textwidth}
 \includegraphics[width=0.7\textwidth]{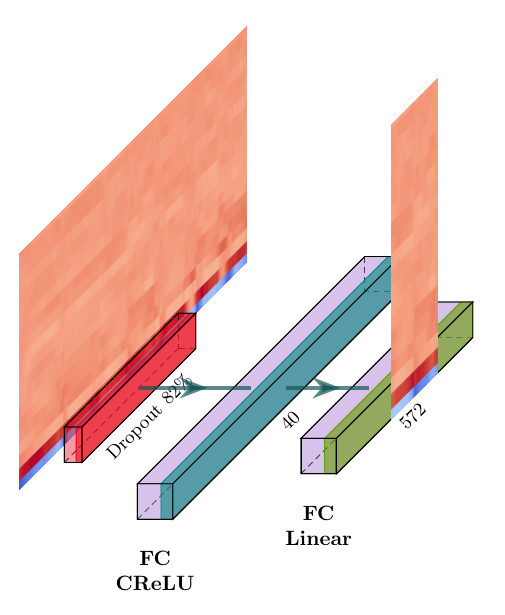}
  \caption{Fully Connected Shallow Neural Network}
  \label{fig:model4}
  \end{minipage}
  \hfill
  \vspace{0.00mm} 
  \begin{minipage}[b]{0.49\textwidth}
 \includegraphics[width=1.2\textwidth]{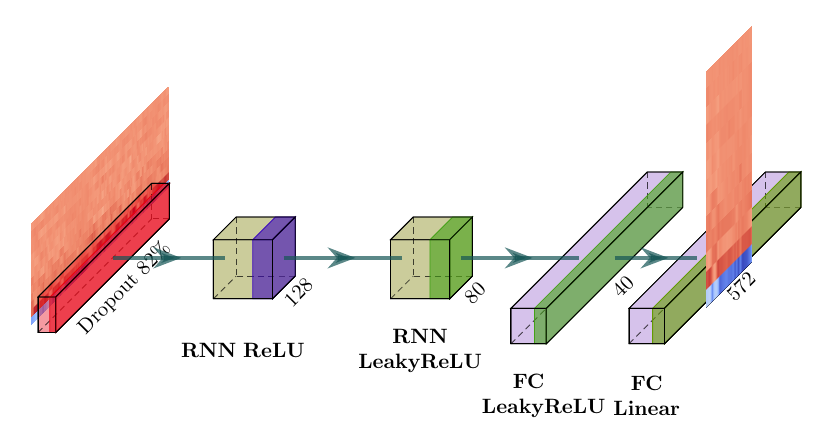}
  \caption{Recurrent Neural Network}
  \label{fig:model7}
  \end{minipage}
\begin{minipage}[b]{1\textwidth}
  \centering
  \includegraphics[width=\textwidth]{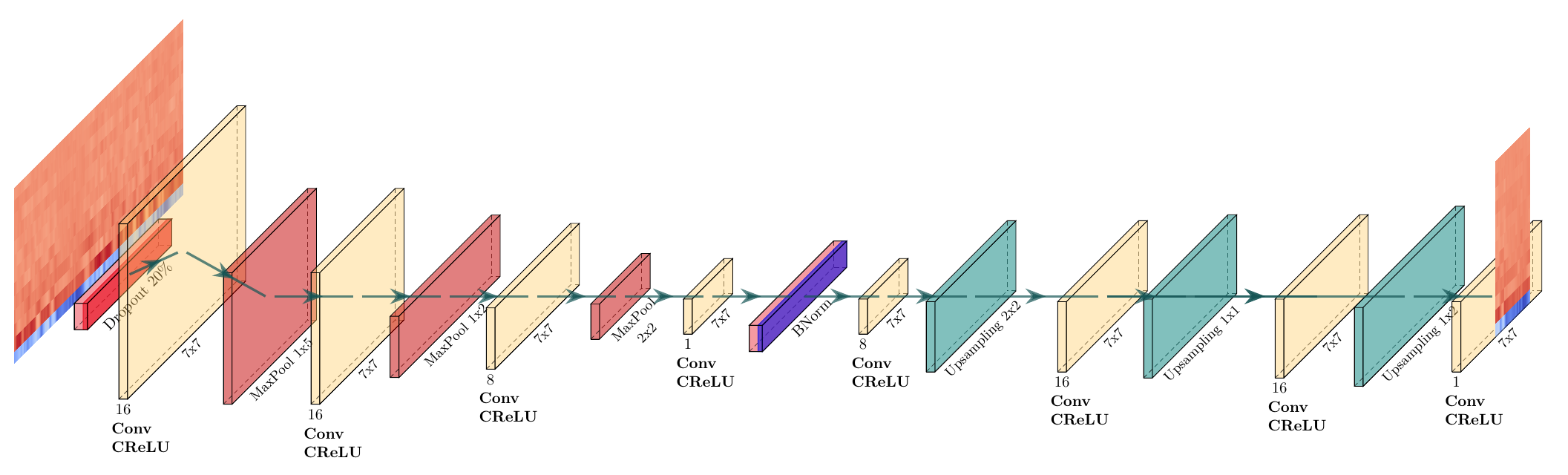}
  \caption{Fully Convolutional Neural Network}
  \label{fig:model6}
\end{minipage}

\end{figure}

To try to find the best topology for the Speech2Phone method, we propose the following experiments:

\begin{itemize}
    \item \textbf{Experiment 1 (Dense)}: this experiment consists of a fully connected feed-forward neural network with one hidden layer for embedding extraction. The architecture of the model used in this experiment is shown in the Figure \ref{fig:model4}.

    \item \textbf{Experiment 2 (Conv)}: This experiment is based on Experiment 1, but it uses a fully convolutional neural network. This network has only convolutional layers, in addition the decoder uses upsampling layer. \textcolor{black}{Following \citep{chung2020in} we use 2D convolutions}. The speech segments can be seen as a bidimensional image matrix where columns are time steps and the rows are cepstral coefficients. Convolutions have the advantage of being translationally invariant in this matrix. As we try to reconstruct a specific phoneme, this is a desired property, since the location of the instance may contain the target phoneme may occur in different parts of the window. 
    The model architecture used in this experiment is shown in the Figure \ref{fig:model6}. 

    \item \textbf{Experiment 3 (Recurrent)}: This experiment is based on Experiment 1, but it consists of a recurrent fully-connected neural network for embedding generation. The five-second window is split into five segments containing one second each, in which the model analyzes one at a time. Considering that the recurrence window is small, we used classical recurrence instead of long term models like LSTM \citep{goodfellow2016}, since vanishing gradients are less prone to occur. If the phoneme of interest happens in one of the five fragments, the recurrence network should store it in its memory before reconstructing it in the final step, potentially improving the reconstruction accuracy. The approach reduces the number of learned parameters and, consequently, also improves training times. The model architecture used in this experiment is shown in the Figure \ref{fig:model7}. 
\end{itemize}


\subsection{Speech2Phone results}\label{sec:results:experiuments:Speech2Phone}

Table \ref{tab:results-Speech2Phone-experiments} shows the results of the Speech2Phone experiments, showing accuracy obtained in the open-set scenario, that is, evaluations using different speakers for training and testing the models and closed-set, where the training and test speakers are the same.

\begin{table}[ht]

\centering
\caption{Results of Speech2Phone Experiments}
\label{tab:results-Speech2Phone-experiments}
\begin{tabular}{l|l|r|r}
\hline
\textbf{Experiment }        & \textbf{Scenario (Subset)}              & \textbf{Accuracy}  & \textbf{Test speech segments} \\\hline
\multirow{2}{*}{\textbf{1 (Dense)}} & closed-set  ($A_2$)  & 77.50    &  100            \\
                   & open-set ($B_1+B_2$) & 76.96    &  1,197          \\
\multirow{2}{*}{\textbf{2 (Conv)}} & closed-set ($A_2$)   & 100.00    &  100            \\
                   & open-set($B_1+B_2$)  & 64.43    &  1,197          \\
\multirow{2}{*}{\textbf{3 (Recurrent)}} & closed-set ($A_2$)   & 88.75    &  100            \\
                   & open-set ($B_1+B_2$) & 50.28    & 1,197         \\ \hline
\end{tabular}

\end{table}
In this set of experiments, the neural models received 5 seconds of audio from a specific speaker and were induced to reconstruct 1 second of the phoneme /a/ in the voice of that speaker. The goal is to obtain good results with little training data in contrast to the GE2E and AngleProto loss functions, which need a large data set for good performance. To conduct the proposed experiments, samples from 20 speakers were used for training and samples from another 20 speakers for testing, as previously discussed in Section \ref{sec:method:Speech2Phone}.

Experiment 1 explored the use of a fully connected neural network and in an open-set scenario, it obtained an accuracy of 76.96\%, the best accuracy in the open-set scenario for all experiments. On the other hand, in the closed-set scenario, it obtained 77.50\% accuracy. This was the worst result for all closed-set experiments. We believe that the fully connected model achieved the best results on the open-set due to the low number of parameters, thus being less prone to overfit. In addition, the dataset is very small and in this way, deeper models are very likely to memorize features dependent on the speaker or noise artifacts. Apparently, the recurrent and convolutional models specialized in extracting particular features for the speakers in the training set in order to reconstruct the output. In this way, as the deeper models learn specific characteristics of the speakers, their generalization for new speakers is impaired, having good performance in the closed-set scenario and a drop in performance in the open-set.

Experiment 2, which explored the use of a fully convolutional neural network for generating embeddings, presented the second best result (64.43\%) in the open-set scenario. On the other hand, in the closed-set, the model achieved the best result obtaining an accuracy of 100\%. 
Translational invariance is a useful feature from convolutional networks and can be used to detect specific phonemes independently of where their occur in the audio. We believe that convolutional models are suitable for this task; however, due to the low amount of data the models learn characteristics dependent on speakers, which leaves this model at a disadvantage in open-set scenario, having a worse performance than Experiment 1.

Experiment 3 explored the use of a recurrent neural network with fully connected layers for generating embeddings, resulting in the worst accuracy in the open-set scenario (50.28\%) and the second worst in the closed-set scenario (88.75\%). Recurrent models can perform a more detailed analysis on the input audio, searching patterns in one input fragment at a time. However, a problem may happen when this pattern is split in a different analysis window. The simple recurrent model tested could not overcome this issue. We also evaluated, in preliminary experiments, a recurrent LSTM network, but it did not perform as well as simple recurrence, as there is no need for long-term memory in a 5-step analysis process.

In the open-set scenario, the superiority of the fully connected model (Experiment 1) is noticeable. This is because they are able to generalize better for new speakers and have proven to be less prone to overfitting for the task. In addition, the fully connected model was able to maintain very close accuracy in the closed-set scenario (77.50\%), compared to the open-set scenario (76.96\%), while the other models had a great drop in performance. On the other hand, the fully convolutional model (Experiment 2) also showed promising results with a performance 12.53\% below the fully connected model. 

\section{Application: Speaker Verification} \label{sec:speaker-verification-experiments}

Section \ref{sec:method:veri} presents the proposed experiments to compare our method with the state of the art in the literature. On the other hand, section \ref{sec:results:experiuments:Speech2Phone} presents the results of speaker verification experiments.

\subsection{Speaker verification Experiments}\label{sec:method:veri}

An important question is how well a model trained with the Speech2Phone method performs in speaker verification. To answer this question we compared Speech2Phone with the Fast ResNet--34 model proposed by \citep{chung2020in} trained using the Angular Prototypical \citep{chung2020in} and GE2E \citep{wan2018generalized} losses function. We chose this model because it presents state-of-the-art results in the VoxCeleb \citep{nagrani2017voxceleb} dataset, in addition the authors made the pre-trained models available in the Github repository\footnote{https://github.com/clovaai/voxceleb\_trainer}. 

To compare the models, we chose two datasets, one of which is multi-language, presented in Section \ref{sec:prepro-veri}. As Speech2Phone receives 22 kHz sample rate audio, while Fast ResNet--34 16 kHz audio input. We cannot directly compare the results of the models in the VoxCeleb test dataset. As voxceleb is sampled at 16 kHz, it would not be a fair comparison to resample audios from 16 kHz to 22 kHz. Therefore, we chose other datasets with a sample rate greater than 22 kHz. In addition, using the dataset test that the Fast ResNet--34 model was trained on could put the Speech2Phone model at a disadvantage.

The audios for each dataset were resampled to 16 kHz for the test with the Fast ResNet--34 model and to 22 kHz for the test with Speech2Phone, thus making a fair comparison between the models. For comparison, we use the metric Equal Error Rate (EER) \citep{cheng2004method}, the lower the EER is, the greater the accuracy becomes. 

To compare the best model trained with the Speech2Phone method with the SOTA we propose the following experiments:
\begin{itemize}
    \item \textbf{Experiment 4 (Speech2Phone)}: This experiment uses the same model and hyperparameters as experiment 1; however, the model was trained with the entire Speech2Phone dataset. Therefore, the model was trained with 2,394 speech segments of 40 speakers totaling approximately 3 hours and 23 minutes of speech.
    \item \textbf{Experiment 5 (Fast AngleProto)}:  This experiment uses the Fast ResNet--34 model, proposed in \citep{chung2020in}, this model is trained with the Angular Prototypical loss function in the Voxceleb dataset, which has approximately 2,000 hours of speech by approximately 7,000 speakers. This model achieves an EER of approximately 2.2 in the voxceleb dataset test set, as reported in \citep{chung2020in}.
    \item \textbf{Experiment 6 (Fast GE2E)}: This experiment also uses the Fast ResNet--34 model, and was also trained with the same number of speakers and hours of experiment 5. However the model is trained using the GE2E loss function. 
\end{itemize}


Unlike the Fast ResNet--34 model, which accepts audios of varying sizes, our model receives an MFCC of just 5 seconds of speech as input. However, in the VCTK and CV datasets we have audios with variable sizes. Therefore, for a fair comparison and each model to have access to all the audio content for calculating the EER of the Speech2Phone model, we proceed as follows. For audios longer than 5 seconds, we use the overlapping technique, as described in Section \ref{sec:prepro-Speech2Phone}. Therefore, after applying the overlapping technique for a six-second audio, two five-second samples are obtained, the resulting embedding for that sample is the average between the predicted embedding for these two five-second samples. On the other hand, for sample less than 5 seconds, we repeat the audio frames until reaching at least 5 seconds, for example, a three-second audio is repeated once, thus obtaining a six-second audio, where the resulting audio applies the overlapping technique, since we audios longer than 5 seconds.

\subsection{Speaker Verification Results}\label{sec:results:experiuments:eer}

In this set of experiments we compared the performance of our best experiment trained with the Speech2Phone method with the Fast ResNet--34 model proposed by \citep{chung2020in}, trained with approximately 2,000 hours of speech using the Angular Prototypical loss function \citep{chung2020in} and GE2E \citep{wan2018generalized}. Table \ref{tab:eerexps} shows the EER of these experiments in English using the VCTK dataset, in Portuguese (PT), Spanish (ES) and Chinese (ZH) using the Common Voice dataset. 

\begin{table}[ht]

\centering
\caption{Results for speaker verification experiments}
\label{tab:eerexps}
\begin{tabular}{llr}
\hline
\textbf{Experiment}                           &\textbf{ Datasets} & \textbf{EER (\%)}     \\\hline
\multirow{5}{*}{\textbf{4 (Speech2Phone)}}   & VCTK     & \textbf{22.7041} \\
                                & CV PT & 13.6805 \\
                                & CV ZH & 10.3909\\
                                & CV ES & 7.7551 \\\hline
\multirow{5}{*}{\textbf{5 (Fast AngleProto)}} & VCTK     & 23.8011 \\
                                & CV PT &  \textbf{7.2468}\\
                                & CV ZH & \textbf{7.2666}\\
                                & CV ES &  \textbf{2.8622}\\\hline
\multirow{5}{*}{\textbf{6 (Fast GE2E)}} & VCTK     & 27.0647 \\
                                & CV PT & 14.0751 \\
                                & CV ZH & 12.9563\\
                                & CV ES &  5.0530\\\hline
                                
                                                                
\end{tabular}

\end{table}

Experiment 4, which consisted of the best Speech2Phone experiments trained using the entire Speech2Phone dataset, that is, approximately 3 hours and 23 minutes of speech from 40 different speakers. This experiment achieved the best EER of all experiments in the VCTK dataset. For the Common Voice dataset, the model achieved the second best result for the PT and ZH subsets, being surpassed only by the Fast ResNet--34 model trained with the Angular Prototypical loss function (Experiment 5). On the other hand, in the ES subset, this experiment had the worst performance of all experiments.

Experiment 5, which consisted of the Fast ResNet--34 model trained with the Angular Prototypical loss function, obtained the second best EER in the VCTK dataset, only surpassed by experiment 4 which was trained with the Speech2Phone method. In addition, this experiment achieved the best result in all subsets of the Common Voice dataset.

Experiment 6, which consisted of the Fast ResNet--34 model trained with the GE2E loss function, obtained the worst EER in the VCTK dataset. This experiment also had the worst EER in the ZH and PT subsets of Common Voice. However, the model achieved the second best EER in the ES subset of common voice.

Experiment 4 showed the potential of the Speech2Phone approach, which despite having been trained with only 3 hours and 23 minutes of speech from only 40 speakers, managed to better results in 3 of the 4 evaluated subsets. Experiment 6, which is the Fast ResNet--34 model trained with 2000 hours of speech and approximately 7,000 speakers using the GE2E loss function. In addition, this experiment was able to perform better than both Experiments 5 and 6 in the VCTK dataset.

Experiment 6 compared to Experiment 4, achieved a better result in 3 of the 4 evaluated subsets, this was already expected due to the difference of more than 1.996 hours of speech and 6.960 speakers between the training datasets of the two models. In addition, the Experiment 4's training dataset has only male voices, as the reconstruction of a female voice is different and the VCTK and Common voice datasets have female speakers this should cause a decrease of performance at test time. Another point is that the training dataset used in Experiment 4 is a high quality dataset that has a low amount of background noise, so the model probably did not learn to ignore noise and the reconstruction of the /a/ phoneme is harmed. Despite this quality and absence of noise facilitating learning during training, the model is impaired in a noisy situation. Another indication of this problem can be seen by its better performance than all the other experiments in the VCTK dataset, which is a high quality and low noise, built for speech synthesis and voice transfer applications. The performance of the model drops, compared to experiment 5, when the model is used in audios recorded in uncontrolled environments such as Commom Voice.

In Experiments 5 and 6 we verified what has already been shown in \citep{chung2020in}, that the Fast ResNet--34 model trained with the AngleProto loss function in the Voxceleb dataset obtains an EER higher than the same model trained with the GE2E loss function. However, the authors in their work compared the models only in the VoxCeleb test dataset, we on the other hand, made a comparison using 4 different languages and different datasets, thus making a broader comparison.

An important consideration is that given the way we propose our experiments, we cannot say that language is a factor that decreases or increases the performance of the models. The high values of EER in the English language, for example, are due to the way the speaker verification dataset was set up. The VCTK has only 109 speakers and many samples were considered for each speaker as can be seen in Table \ref{tab:preproveri}. A greater number of test instances make the task more difficult and tend to increase the EER values. Therefore, we can only compare the individual performance of each model in the datasets and we cannot discuss decrease or increase in performance with the language change.

\section{Conclusions and future work} \label{sec:conc}

In this article, we proposed a novel training method for speaker recognition models, called Speech2Phone. To enable the training of this method, we also built a novel dataset. Fully connected, fully convolutional and recurring models were explored. We observed that the fully connected models have a better performance in open-set scenarios, although they have had the worst performance for the closed-set scenario, while convolutional models have a better performance in the closed-set scenario, but they do not generalize well for the open-set scenario. The best model in our experiments was trained on 3 hours and 23 minutes of speech and compared to two SOTA models in the VoxCeleb dataset, which were trained with approximately 2000 hours of speech. The results obtained were comparable to SOTA even with an amount of data approximately 500 times smaller.

This work contributes directly to the area of speaker recognition, presenting a promising method for training speaker recognition models. In addition, the model proposed here can be used in tasks such as speech synthesis \citep{ping2017deep}, voice cloning \citep{arik2018neural} and multilingual speech conversion \citep{zhou2019cross}. In these tasks, the speaker recognition system embeddings are used to represent the speaker. In addition, an advantage of Speech2Phone in relation to the models proposed in \citep{chung2020in} is the speed of execution, as our best model is a fully connected shallow neural network, making it faster. This feature is very desirable for applications due to the need to run in real time.



As our model demands a specific dataset format, we were not able to evaluate its training in a large dataset. We plan to address this issue in future works.  For this we intend to increase the model's training dataset as much as possible, and make it public. Additionally, we intend to explore the use of multispeaker speech synthesis \citep{cooper2020zero} and voice cloning \citep{arik2018neural} to generate a dataset with more speakers and a larger vocabulary. On the other hand, we intend to investigate the possibility of a hybrid method that uses a Speech2Phone technique and 
Angular Prototypical loss function, thus being able to learn from a larger amount of data and at the same time guide the model's learning with the reconstruction of a phoneme. In addition, we intend to explore the insertion of noise in the training dataset in order to make the model more robust to noise. 




\bibliographystyle{splncs03}
\bibliography{paper}

\end{document}